\documentclass[10pt,twocolumn,letterpaper]{article}

\usepackage{iccv}
\usepackage{times}
\usepackage{epsfig}
\usepackage{graphicx}
\usepackage{amsmath}
\usepackage{amssymb}
\usepackage{authblk}
\usepackage{algorithm}
\usepackage{algorithmic}
\usepackage[colorlinks,linkcolor=blue,bookmarks=false]{hyperref}

\iccvfinalcopy 

\def\iccvPaperID{****} 

\ificcvfinal\fi

\begin{document}


\title{GradNet: Gradient-Guided Network for Visual Object Tracking}

\author{Peixia Li\textsuperscript{\dag}, Boyu Chen\textsuperscript{\dag}, Wanli Ouyang\textsuperscript{\S}, Dong Wang\textsuperscript{\dag}\thanks{Corresponding Author: Dr. Dong Wang}, Xiaoyun Yang\textsuperscript{\ddag},
Huchuan Lu\textsuperscript{\dag} \\
{\dag} Dalian University of Technology, China,
{\S} The University of Sydney, Australia, \\
{\ddag} China Science IntelliCloud Technology Co., Ltd \\
{\sl{\small{\{pxli, bychen\}@mail.dlut.edu.cn, wanli.ouyang@sydney.edu.au, \{wdice,lhchuan\}@dlut.edu.cn, xiaoyun.yang@intellicloud.ai}}}
}



\maketitle
\ificcvfinal\thispagestyle{empty}\fi

\begin{abstract}
	The fully-convolutional siamese network based on template matching has shown great potentials in visual tracking.
	During testing, the template is fixed with the initial target feature and the performance totally relies on the general matching ability of the siamese network.
	However, this manner cannot capture the temporal variations of targets or background clutter.
	In this work, we propose a novel gradient-guided network to exploit the discriminative information in gradients and update the template in the siamese network through feed-forward and backward operations.	
	%
	%
	To be specific, the algorithm can utilize the information from the gradient to update the template in the current frame.
	In addition, a template generalization training method is proposed to better use gradient information and avoid overfitting.
	To our knowledge, this work is the first attempt to exploit the information in the gradient for template update in siamese-based trackers. Extensive experiments on recent benchmarks demonstrate that our method achieves better performance than other state-of-the-art trackers.
	The source codes are available at \href{https://github.com/LPXTT/GradNet-Pytorch}{https://github.com/LPXTT/GradNet-Pytorch}.
\end{abstract}

\section{Introduction}
Visual object tracking is an important topic in computer vision, where the target object is identified in the initial video frame and successively tracked in subsequent frames.
%
%
In recent years, deep networks~\cite{Wang-CVPR16-STCT,Bertinetto-ECCV16-SiamesFC,Huang-CoRR17-east,zhang-eccv2018-structsiam,Li-PR18-survey, iccv19_SPLT}
have significantly improved the tracking performance due to their representation prowess.
%

There are two groups of deep-learning-based trackers. The first group \cite{Wang-ICCV15-FCNT,Nam-CVPR16-MDNet,crest-iccv2017,Chen-pr2019-mam} improves the discriminative ability of deep networks
by frequent online update. They utilize the first frame to initialize the model and update it every few frames. Timely online update enables trackers to capture target variations but also requires more computational time.
Therefore, the speed of these trackers generally cannot meet the real-time requirements.

\begin{table*}[t]
	\centering
	\caption{The number of backward iterations to update the template of SiameseFC. `LR' means learning rate; `$n\times$' means $n$ times the basic learning rate;
	`ITERs' means the needed iterations to converge. There is no proper step to converge by one iteration.}
	\centering
	\setlength{\tabcolsep}{1.9mm}{
		\begin{tabular}{l||c|c|c|c|c|c|c|c|c|c|c|c|c|c|c}
			\hline
			{\textbf{LR} } & {1$\times$} & {3$\times$} & {5$\times$}& {7$\times$}& {9$\times$}& {10$\times$}& {30$\times$}& {50$\times$}& {70$\times$}& {90$\times$}& {100$\times$}& {500$\times$}& {1000$\times$}& {3000$\times$}& {5000$\times$}\\
			\hline
			{\textbf{ITERs}} & 449 & 136 & 77 & 64 & 60 & 58 & 59 & 51 & 54 & 56 & 55 & 54 & 61 & 67 & $\infty$ \\
			\hline
		\end{tabular}}
		\label{table:step}
		\vspace{-2mm}
	\end{table*}

\begin{figure}[t]
	\centering
	\includegraphics[width=1\linewidth]{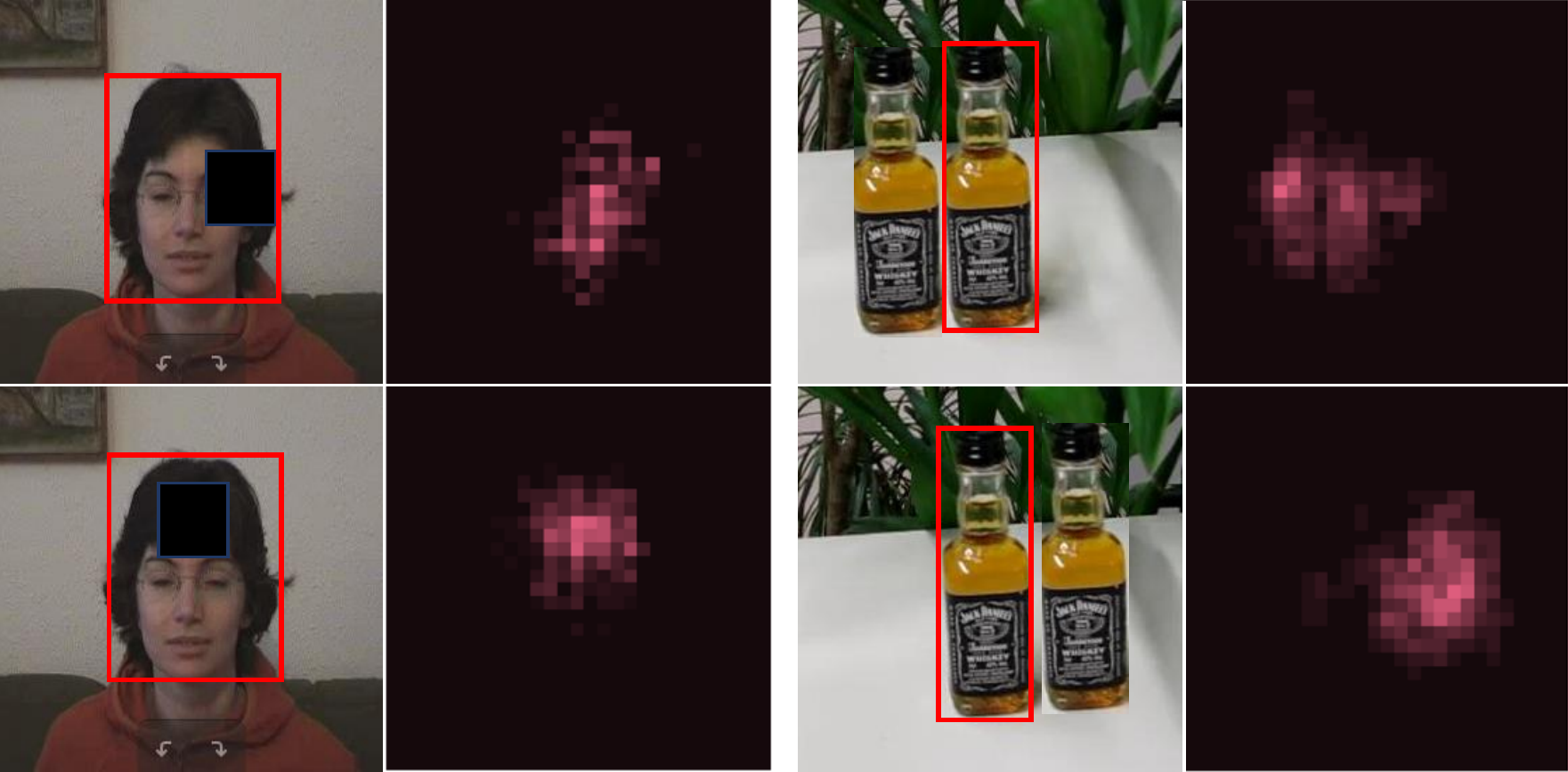}\\
	\caption{The motivation of our algorithm. Images in the first and third column are target patches in SiameseFC. The other images show absolute values of their gradients, where the red regions have large gradient. As we can see, the gradient values can reflect the target variations and background clutter.} \label{fig:m1}
\end{figure}

Siamese-based trackers are representative in the second group~\cite{Bertinetto-ECCV16-SiamesFC, zhang-eccv2018-structsiam,Li-2018CVPR-siameserpn} which is totally based on offline training. They learn the similarity between objects in different frames through massive offline training.
During online testing, the initial target feature is regarded as template and used to search the target in the following frames. These methods need no online updating, thus, they usually run at real-time speeds.
However, these methods cannot adapt to appearance variations of target without important online adaptability, thereby increasing the risk of tracking drift.
To solve this problem, many researches ~\cite{Guo-dsiam-iccv2017, zhu-flowtracker-eccv18, Yang-memtracker-eccv2018} present different mechanisms to update template features. However, these methods only focus on combining the previous target features, ignoring the discriminative information in background clutter. This results in a big accuracy gap between the siamese-based trackers and those with online update.

Generally, gradients are calculated through the final loss which considers both positive and negative candidates. Thus, gradients contain the discriminative information to reflect the target variations and distinguish the target from background clutter. As shown in Figure~\ref{fig:m1}, when objects are occluded with noise or similar objects coexist at the neighborhood of the target, the absolute value of gradients at these locations are prone to be higher. The high value in gradients can force the template to focus on these regions and capture the core discriminative information. Most gradient-based trackers~\cite{Wang-ICCV15-FCNT, crest-iccv2017} concentrate on hand-designed optimization algorithms, such as momentum~\cite{momentum}, Adagrad~\cite{adagrad}, ADAM~\cite{adam} and so on. These algorithms need hundreds of iterations to converge, which lead to more computation and a lower speed.
How to take a trade-off between the speed and accuracy of update is still a problem.


If we expect to reduce the number of training iterations but still keep online update through gradients, the extreme case is to adapt the template through one backward propagation.
However, training by one backward propagation is a difficult task.
As shown in Table~\ref{table:step}, there is no proper learning rate to make the template of SiameseFC converge through one iteration.
%
Generally, even with the optimal step length, moving according to the gradient at only one iteration cannot update the template properly, because the normal gradient-based optimization is a nonlinear process.
On the other hand, we can learn a nonlinear function by CNNs, which simulates the non-linear gradient-based optimization by exploring the rich information in gradients.
Therefore, we propose a gradient-guided network (GradNet) to perform gradient-guided adaptation in visual tracking.
The GradNet integrates the adaptation process that consists of two feed-forward and one backward calculation, simplifying the process of gradient-based optimization.

\begin{figure}[t]
	\centering
	\includegraphics[width=1\linewidth]{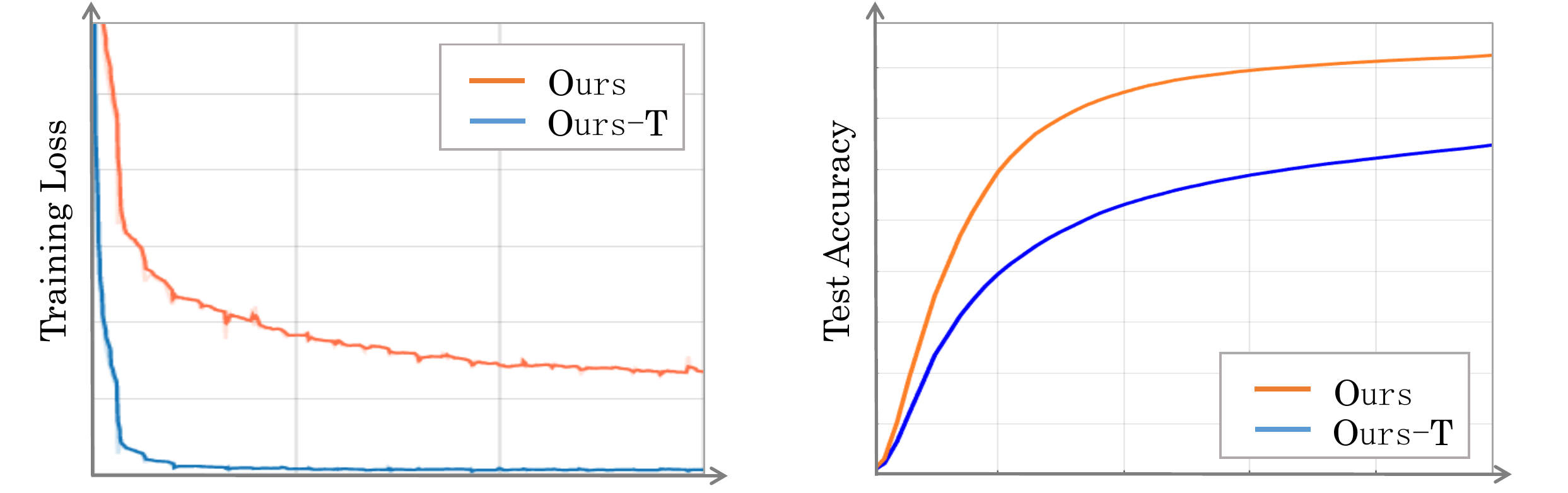}\\
	\caption{The training and testing plots of models through normal training (Ours-T) and our training method (Ours). The left map shows the error between the predicted map and the real map during training and the right map shows the accuracy during testing.} \label{fig:m_l}
\end{figure}

It is a very tough task to train a robust GradNet due to two main reasons.
%
The first reason is that the network is prone to use the appearance of the template instead of using the gradient for tracking (details can be found in Section~\ref{sec::training}), because learning to use the gradients is more difficult than learning to use appearance.
%
The second reason is that the network is prone to overfit. As shown in Figure~\ref{fig:m_l}, the model with normal training (Ours-T) can quickly get a low distance error but its test accuracy is not promising, compared with our model.
%
To handle these issues, we propose a template generalization method to effectively explore gradient information and avoid overfitting.

The major contributions can be summarized as follows:
\begin{itemize}
	\setlength{\itemsep}{4pt}
	\setlength{\parsep}{4pt}
	\setlength{\parskip}{4pt}
	\item A GradNet is proposed to conduct gradient-guided template updating for visual tracking.
	\item A template generalization method is proposed to ensure strong adaptation ability and avoid overfitting.
	\item Extensive experiments conducted on four popular benchmarks show that the proposed tracker achieves promising results at a real-time speed of $\emph{\textbf{80fps}}$.
\end{itemize}

\section{Related Work}
\subsection{Siamese Network based Tracking}
SiameseFC~\cite{Bertinetto-ECCV16-SiamesFC} is the most representative trackers based on template matching.
Bertinetto~\emph{et al.}~\cite{Bertinetto-ECCV16-SiamesFC} present a siamese network
with two shared branches to extract features of both the target and the search region.
%
%
%
During online tracking, the template is fixed as the initial target feature and the tracking performance mainly relies on the discriminative ability of the offline-trained network.
Without online updating, the tracker achieves beyond real-time speed.
Similarly, SINT~\cite{Tao-CVPR16-SINT} also designs a network to match the initial target with candidates in a new frame. Its speed is much lower because hundreds of candidate patches are sent into the network instead of one search image.
Another siamese-based tracker is GOTURN~\cite{Held-ECCV16-GOTURN} which proposes a siamese network to regress the target bounding box with a speed of 100fps.
All these methods are lack of important online updating.
The fixed model cannot adapt to appearance variations, which makes the tracker easily disturbed by similar instances or background noise.
In this paper, we choose SiameseFC as our basic model and propose a gradient-guided method to update the template.

\subsection{Model Updating in Tracking}
Timely updating is essential to keep trackers robust. There are three main dominant strategies of model updating, including template combination, gradient-descent based and correlation-based strategies.

\vspace{-4pt}
{\flushleft {\bf{Template Combination. }}}
Algorithms~\cite{Guo-dsiam-iccv2017, zhu-flowtracker-eccv18} based on template combination aim to effectively combine the target features from previous frames.
%
%
Guo~\emph{et al.}~\cite{Guo-dsiam-iccv2017} propose a fast transformation learning model
to enable effective online learning from previous frames.
Zhu~\emph{et al.}~\cite{zhu-flowtracker-eccv18} utilize the optical flow information to convert templates and integrate them according to their weights.
All these methods focus on using the information of templates, which ignore the background clutter.
Different from these methods, we take full use of the discriminative information in backward gradients
instead of just integrating previous templates.

\vspace{-4pt}
{\flushleft {\bf{Gradient-descent based approaches. }}}
Deep trackers~\cite{Wang-ICCV15-FCNT, crest-iccv2017} based on gradient descent explore the discriminative information in backward gradients to update the model through hundreds of iterations.
Wang~\emph{et al.}~\cite{Wang-ICCV15-FCNT} train two separate convolutional layers to regress Gaussian maps with the initial frame and update these layers every few frames.
Similarly, Song~\emph{et al.}~\cite{crest-iccv2017} also utilize a number of gradient descent iterations in initialization and online update procedures.
These trackers need many training iterations to capture the appearance variations of the target, which makes the tracker less effective and far from real-time requirements.
We propose a GradNet that needs only one backward propagation and two forward propagations to update the template effectively. Besides, our template generalization method for handling overfitting is not investigated in existing works.

\vspace{-4pt}
{\flushleft {\bf{Correlation based Tracking. }}}
Correlation based trackers~\cite{Henriques-TPAMI15-KCF, Ma-ICCV15-HCFT, Danelljan-ECCV16-C-COT,MCPF_TPAMI19,CPF_TIP19,ASRCF} train classifier through circular convolution, which can be quickly calculated in Fourier domain.
The final classifier is trained and updated by solving the closed-form solution of the optimization function.
The classifier training cannot be simulated totally by deep networks, so most correlation based trackers just utilize deep networks to extract robust features.
Differently, our method aims to update the template in an end-to-end network.

\subsection{Gradient Exploiting}
Currently, most deep neural networks adopt gradients in offline training based on hand-designed optimization strategies, such as momentum~\cite{momentum}, Adagrad~\cite{adagrad}, ADAM~\cite{adam} and so on.
These methods usually need expensive computation and large-scale data sets.
How to accelerate the training of deep networks is a hot topic in computer vision.

\vspace{-4pt}
{\flushleft {\bf{Meta Learning. }}}
Meta learning approaches can be broadly divided into different categories, including
optimization-based methods~\cite{ltl-nip16}, memory-based methods~\cite{mann-icml16}, variable-based methods~\cite{mamm-icml17, Andrei-corr18-latent, Li-corr17-metasgd} and so on.
Our algorithm can be seen as an improved version of the optimization-based method~\cite{ltl-nip16} to adapt to the update task in visual tracking. Our approach has three main differences compared with \cite{ltl-nip16}. First, ours only learns to update template, but not the network branch of search region. This is specifically designed for the tracking task. Second, our update process only contains one iteration instead of multiple iterations. Finally, our training of the optimizer includes second-order gradient which is not used in \cite{ltl-nip16}.

\vspace{-4pt}
{\flushleft {\bf{Meta Learning for Tracking. }}}
Despite the popularity of meta learning in many fields, there are few works~\cite{Yang-memtracker-eccv2018,Eunbyung-eccv18-meta-tracker} applying it to visual tracking.
Yang~\emph{et al.}~\cite{Yang-memtracker-eccv2018} design a memory structure to
dynamically write and read previous templates for model updating. Differently, we focus on exploring the discriminative information of gradients.
Eunbyung~\emph{et al.}~\cite{Eunbyung-eccv18-meta-tracker} train the initialization parameters of filters with pixel-wise learning rate offline and utilize a matrix multiplication to update the filters. The update is a linear process. While, our template update is a non-linear process with convolutional layers and Relu. Besides, we use the target feature as the prior information to speed up the update process by providing a good initial value.

\begin{figure*}[t]
	\centering
	\includegraphics[width=1\linewidth]{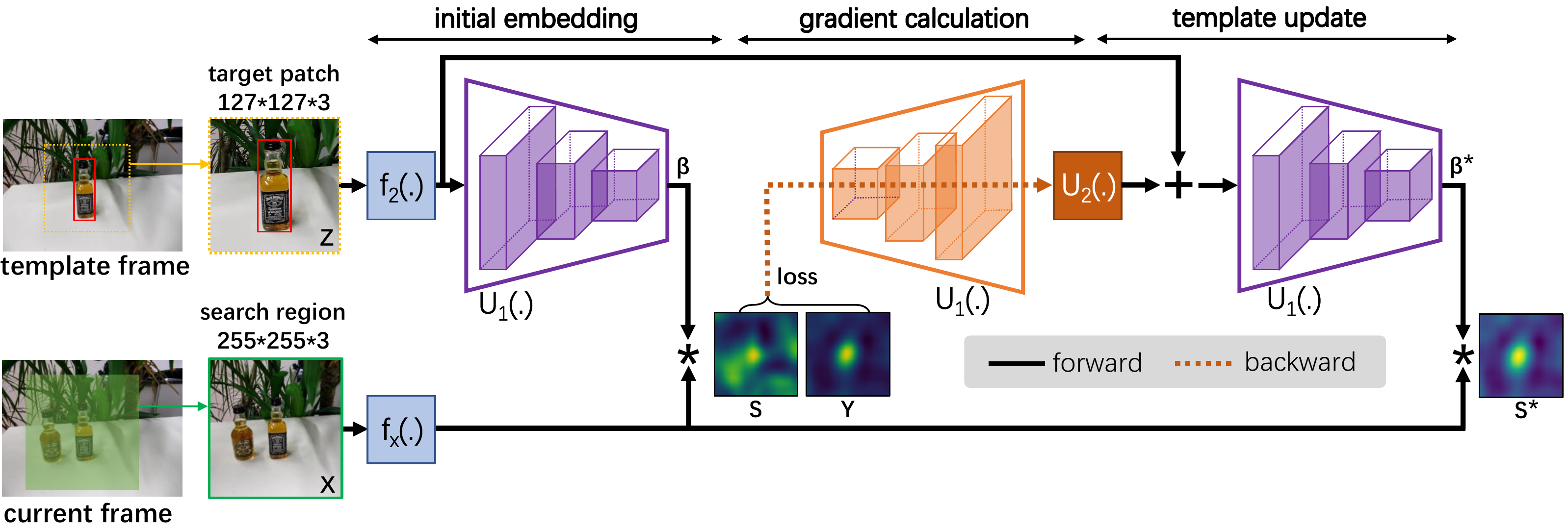}\\
	\caption{The pipeline of the proposed algorithm, which consists of two branches. The bottom branch extracts the feature of search region $\mathbf{X}$ and the top branch (named update branch) is responsible for template generation. The two purple trapezoids in the figure represent sub-nets with shared parameters; the solid and dotted line represents forward and backward propagation respectively.} \label{figs1}
\end{figure*}

\section{Proposed Algorithm}
The whole pipeline of GradNet is shown in Figure~\ref{figs1}, which consists of two branches. One branch extracts features of the search region $\mathbf{X}$ and the other branch generates the template according to the target information and gradients, detailed in Section~\ref{sec::optimizer}.
%
The template generation process consists of initial embedding, gradient calculation and template updating.
First, the shallow target feature $f_2(\mathbf{Z})$ is sent to one sub-net $\mathbf{U}_1$ (shown in purple in Figure~\ref{figs1}) to obtain an initial template $\beta$ which is used to calculate the initial loss $L$.
Second, the gradient of the shallow target feature is calculated through backward propagation, and sent to the other sub-net $\mathbf{U}_2$ (shown in orange in Figure~\ref{figs1}) for being non-linearly converted to better gradient representation.
Finally, the converted gradient is added to the shallow target feature to get an updated target feature which is sent to the sub-net $\mathbf{U}_1$ again to output the optimal template.
It should be noted that the two sub-nets in the initial embedding and template update process share parameters.
The optimal template is used to search targets on search regions through cross correlation convolution.

\subsection{Basic Tracker}\label{sec::tracker}

We adopt SiameseFC~\cite{Bertinetto-ECCV16-SiamesFC} as the basic tracker. $f_x(.)$ is used to model the feature extraction branch for search region, $f_z(.)$ is used to model the feature extraction branch for target region.
We assume that the movement of the target is smooth between two consecutive frames. Thus,
we can crop a search region $\mathbf{X}$ which is larger than the target patch $\mathbf{Z}$ in the current frame, centered at the target's position in the last frame. The final score map is calculated by:
\begin{equation}\label{equ1}
\mathbf{S} = \beta * f_x(\mathbf{X}),
\end{equation}
where $\beta$ is the template to perform an exhaustive search over the search region $\mathbf{X}$, $*$ means cross correlation convolution, $\mathbf{S}$ denotes the score map to find the target.
%
%
In SiameseFC, the template $\beta$ is defined as the deep target feature:
\begin{equation}\label{sia}
\beta_{sia}=f_z(\mathbf{Z}),
\end{equation}
where $\mathbf{Z}$ is the target patch in the first frame.
In order to improve the discriminative ability of the template $\beta$ during online tracking, we design the update branch $U(\alpha)$ to explore the rich information in gradients:
\begin{equation}\label{our}
\beta_{our}=U(\mathbf{Z},\mathbf{X},\alpha),
\end{equation}
where $\alpha$ is the parameter of the update branch
which can not only capture the template information in $\mathbf{Z}$ but also  the background information in $\mathbf{X}$ through gradients.

%

\subsection{Template Generation}\label{sec::optimizer}


%
{\flushleft {\bf{Initial Embedding. }}}
Given the image pair $(\mathbf{X}, \mathbf{Z})$, we want to get the optimal template $\beta^{\star}$ which is suitable to distinguish the target from the background in search region $\mathbf{X}$. First, we get the target feature $f_2(\mathbf{Z})$ (using two convolutional layers) and sent $f_2(\mathbf{Z})$ to the sub-net $U_1$ to get the initial template $\beta$:
\begin{equation}\label{equof2}
\beta=U_1(f_2(\mathbf{Z}),\alpha_1),
\end{equation}
where $\alpha_1$ is the parameter of $U_1$.
The initial template only contains template information without background information. Thus, we need to explore the discriminative information in gradient to make it more robust.
After getting $\beta$, the initial score map $\mathbf{S}$ is calculated through equation (\ref{equ1}).
\vspace{-3pt}
{\flushleft {\bf{Gradient Calculation. }}}
Based on the initial score map $\mathbf{S}$ and the training label $\mathbf{Y}$, we can get the initial loss $L$ by:
\begin{equation}\label{equ:lsx}
L={l\left({\mathbf{S},{{\mathbf{Y}}}}\right)},
\end{equation}
where $l(.)$ is logistic loss function. We utilize this loss to calculate the gradient of $f_2(\mathbf{Z})$ and added it to $f_2(\mathbf{Z})$.
%
Then, the updated target feature is obtained by:
\begin{equation}\label{equ:f2star}
{{h}_{{2}}}(\mathbf{Z})={{f}_{{2}}}(\mathbf{Z}) + U_2(\frac{{\partial}L}{{\partial}{{f}_{{2}}}(\mathbf{Z})}, \alpha_2),
\end{equation}
where $\alpha_2$ is the parameter of $U_2$.
\emph{Here, the gradient is related to $U_1$ and used as the input of the sub-net $U_2$ to calculate the final loss, so the second-order guidance is introduced in the parameter training of the sub-net $U_1$.}

\vspace{-3pt}
{\flushleft {\bf{Template Update. }}}
Finally, we send the updated target feature ${{h}_{{2}}}(\mathbf{Z})$ to
the sub-net $U_1$ again to obtain the optimal template $\beta^{\star}$ and the final score map $S^{{\star}}$ by:
\begin{equation}\label{equ:final}
\begin{gathered}
\beta^{\star}=U_1(h_2(\mathbf{Z}),\alpha_1), \\
\mathbf{S^{\star}}=\beta^{\star}*f_x(\mathbf{X}).
\end{gathered}
\end{equation}
The optimal score map $\mathbf{S^{\star}}$ is utilized to estimate the target position.
Our goal is to let $\mathbf{S^{\star}}$ have the highest value at the target position and lower values at other positions.
Thus, we utilize the loss which is calculated by $\mathbf{S^{\star}}$ to train the update branch:
\begin{equation}\label{equ:argmin}
{{\arg\min}_{{{\alpha}}} \sum{l\left({\mathbf{S^{\star}},{{\mathbf{Y}}}}\right)}}.
\end{equation}

To our knowledge, this work is the first attempt to exploit the discriminative information of gradients to update the template in SiameseFC.
To simplify the introduction of template generation process, we just utilize one image pair here. In the next subsection, we will discuss the training method more generally and detailedly.

\begin{figure}[t]
	\footnotesize
	\begin{center}
		\includegraphics[width=0.99\linewidth]{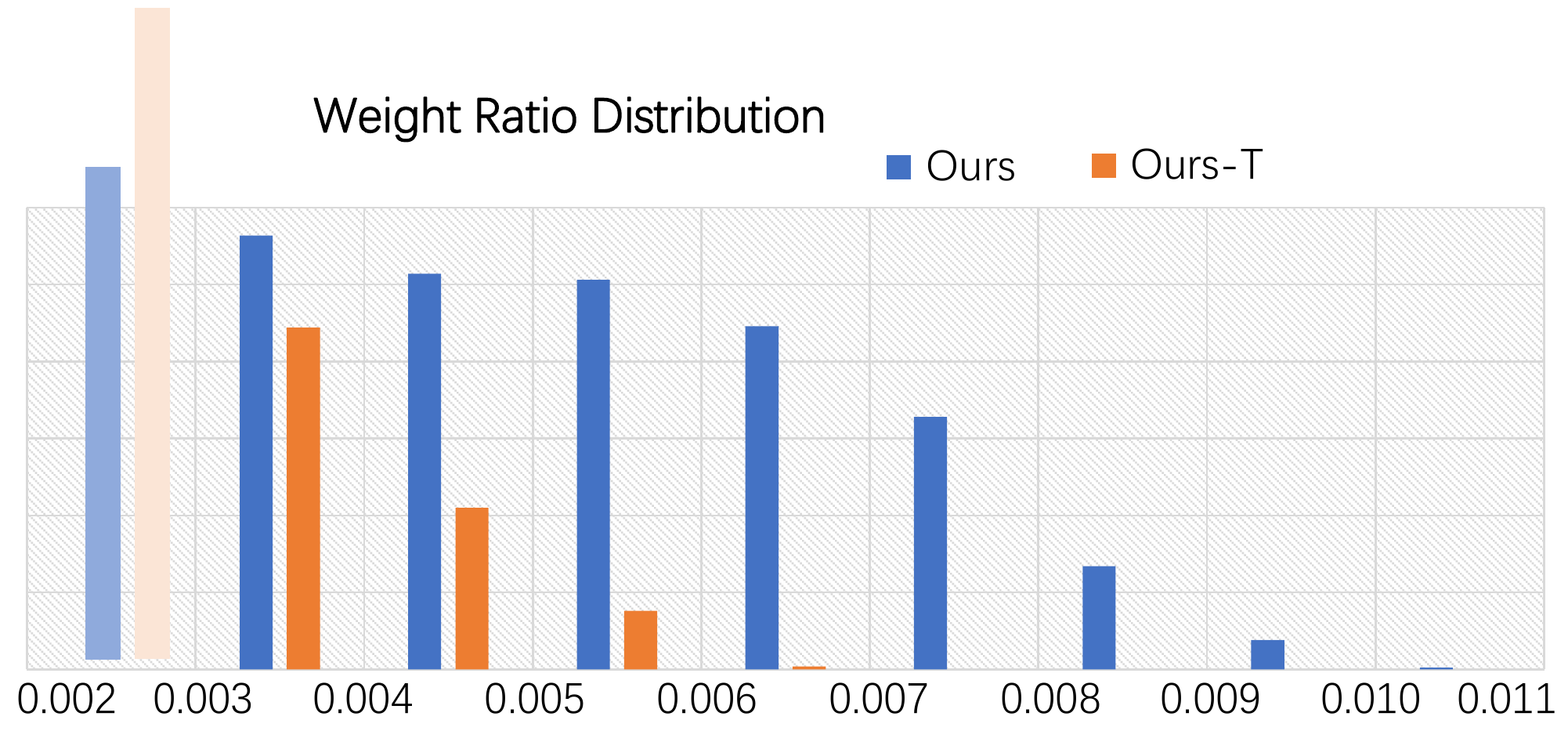}   \\
	\end{center}
	\vspace{-3mm}
	\caption{The distribution of weight ratio between gradients and features. The weight ration is calculated by the absolute value of $\alpha_2$, which reflects the proportion of the gradient during the template update. The rectangles at different positions represent the number of points in those ranges.}
	\label{fig:g-zhu}
\end{figure}

\subsection{Template Generalization}\label{sec::training}
\noindent
{{\bf{Problem of Basic Optimization. }}}
Image pairs from different videos and their training labels form the training set $T=\{({\mathbf{X}}_{{1}},{\mathbf{Z}}_{{1}},{\mathbf{Y}}_{{1}}),({\mathbf{X}}_{{2}},{\mathbf{Z}}_{{2}},{\mathbf{Y}}_{{2}}),\hdots ,({\mathbf{X}}_{{n}},{\mathbf{Z}}_{{n}},{\mathbf{Y}}_{{n}})\}$, $\mathbf{X}_i$ is search region which is larger than target patch $\mathbf{Z}_i$, $\mathbf{Y}_i$ is training label and $n$ is the number of training samples. It should be noted that ${\mathbf{X}}_{{i}}$ and ${\mathbf{Z}}_{{i}}$ are from different frames of the same video, while ${\mathbf{X}}_{{i}}$ and ${\mathbf{X}}_{{j}}$ ($i \neq j$) are from different videos. One simple idea to train our network is to utilize image pairs $(\mathbf{X}_i,\mathbf{Z}_i, \mathbf{Y}_i)$ in the training set $T$ to get optimal template $\beta^{\star}_{i}$ and final score maps $\mathbf{S}^{\star}_i$ by equations (\ref{equof2}$-$\ref{equ:final}).
The update branch is trained through:
\begin{equation}\label{equ:10}
\begin{aligned}
{{\arg\min}_{{{\alpha}}}\sum _{i=1}^{n}{l\left({\mathbf{S}^{\star}_i,{\mathbf{Y}_{i}}}\right)}}.
\end{aligned}
\end{equation}
This method has two main problems according to our experiment. The first one is that the update branch of the network is prone to focus on the template appearance instead of the gradient, because learning to use the gradient is harder than modeling the similarity metric. As shown in Figure~\ref{fig:g-zhu}, the network trained without template generalization has lower weight ratio of gradients. This means that the network focuses less on gradients. The second one is that the network cannot avoid overfitting under this training process as shown in Figure~\ref{fig:m_l}.

\begin{figure}[t]
	\centering
	\includegraphics[width=1\linewidth]{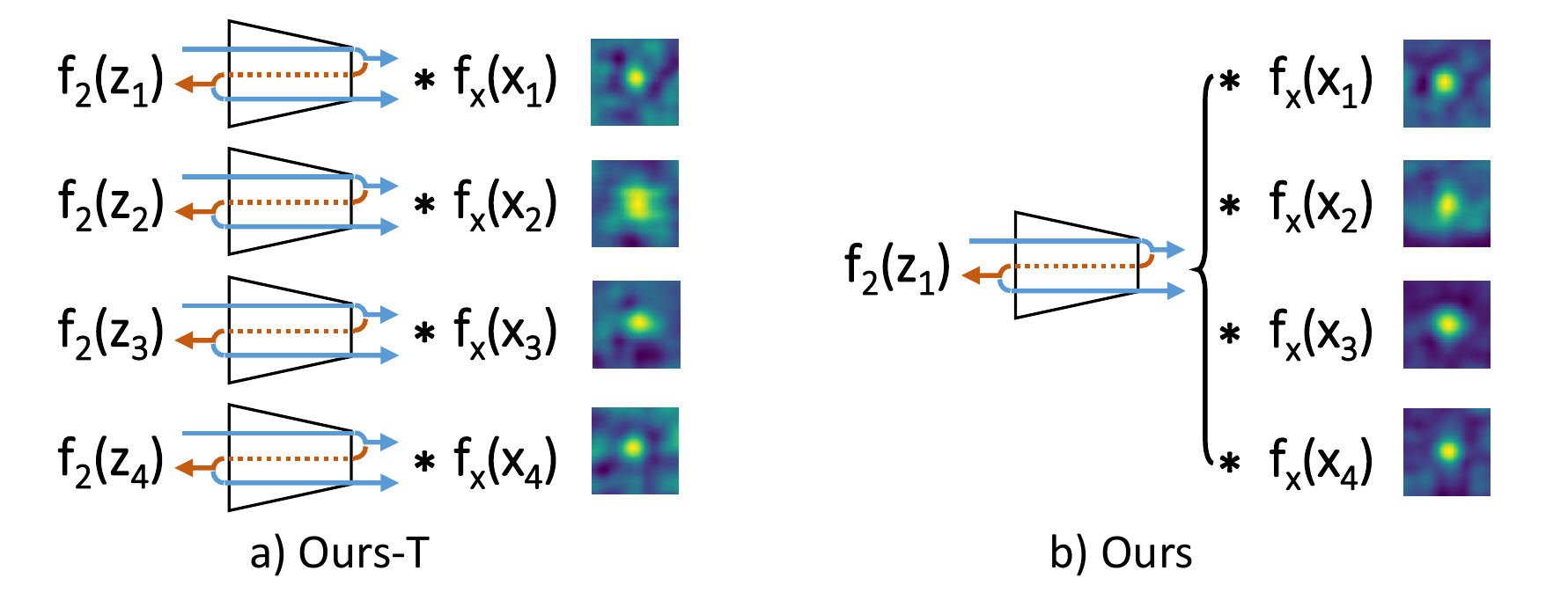}\\
	\caption{ Illustration of `Ours-T' and `Ours' on exploiting templates. `Ours-T' denotes training without template generalization; `Ours' represents training through template generalization.} \label{fig:train}
\end{figure}

\vspace{4pt}
\noindent
{{\bf{Template Generalization. }}}
Our goal is forcing the update branch to focus on gradients and avoiding overfitting.
Based on these requirements, we propose a template generalization method which adopts search regions from different videos to obtain a versatile template and make it perform well on all search regions in each training batch.
We show the training process of our model without template generalization $(a)$ and our model with template generalization $(b)$ in Figure~\ref{fig:train} based on four image pairs. The main difference is that we utilize one template (instead of four templates) to search targets on four images from different videos.

We choose $k$ ($k=4$ in Figure~\ref{fig:train}) training image pairs from the training set $T$ to form a training batch and utilize the target patch $\mathbf{Z}_1$ in the first image pair to calculate the target feature $f_2(\mathbf{Z}_1)$. The initial template $\beta_1$ can be obtained by equation (\ref{equof2}). Here, $\beta_1$ means the template which is calculated through $\mathbf{Z}_1$.
Then, we utilize $\beta_1$ to find the target on all search regions:
\begin{equation}\label{equ:m1}
\mathbf{S}_i=\beta_1*f_x(\mathbf{X}_i), i=1,2,...k.
\end{equation}
Then, we can obtain the initial loss by equation (\ref{equ:lsx}) and update the template $\beta_1$ through equations (\ref{equ:f2star}, \ref{equ:final}). After obtaining the updated template $\beta^{\star}_1$, we utilize it to search the target in all search regions $(\mathbf{X}_1, \mathbf{X}_2, ..., \mathbf{X}_k)$ and train the update branch through equation (\ref{equ:10}).
In this way, the $\beta^{\star}_1$ is required to track the targets in $\mathbf{X}_1, \mathbf{X}_2, ..., \mathbf{X}_k$ simultaneously.
To clarify, we show the details in Algorithm~\ref{alg:offline}.

The template generalization offers the target feature with multiple search regions and aims to obtain a general template feature which performs well on all search regions. This strategy can force the network to focus on the gradients during offline training, because the initial target features are misaligned and the gradients are aligned. The sub-nets $U_1$ and $U_2$ need to correct the initial misaligned template according to the gradients and thereby obtaining a great power to update templates according to gradients.
As shown in Figures~\ref{fig:m_l} and~\ref{fig:g-zhu}, the template generalization algorithm can effectively avoid overfitting and pay attention on gradients.

\begin{algorithm}[h]
	\caption{Offline training the update branch}
	\label{alg:offline}
	{\textbf{Input: }}Training samples $(\mathbf{I}_{{1}},\mathbf{I}_{{2}},\hdots,\mathbf{I}_{{n}})$ from different videos and gaussian maps $(\mathbf{Y}_{{1}},\mathbf{Y}_{{2}},\hdots,\mathbf{Y}_{{n}})$\\
	{\textbf{Output: }}Trained weights ${{\alpha}}$ for the update branch.\\
	\begin{algorithmic}
	\vspace{-2mm}
		\STATE {Initialize the update branch with weights
			${{\alpha}^{0}}$.\\
			Initialize the feature extraction part of the tracker with parameters from SiameseFC~\cite{Bertinetto-ECCV16-SiamesFC}}.
		
		\STATE Crop template images $\mathbf{Z}$ and search regions $\mathbf{X}$ from the training samples to construct the training set  $T=\{(\mathbf{X}_{{1}},\mathbf{Z}_{{1}},\mathbf{Y}_{{1}}),(\mathbf{X}_{{2}},\mathbf{Z}_{{2}},\mathbf{Y}_{{2}}),\hdots,(\mathbf{X}_{{n}},\mathbf{Z}_{{n}},\mathbf{Y}_{{n}})\}$.
		
		\WHILE{ not converged}
		\STATE 1. Randomly select $k$ training samples from $T$.
		\STATE 2. Utilize the update branch to get $\beta_1$ and $\beta^{{\star}}_1$.
		\FOR{$i \in {0,\hdots,k}$}
		\STATE (a). $\beta_{1}={{U_1}}({{f}_{{2}}}(\mathbf{Z}_{{1}}),
		{{\alpha}_{{1}}})$
		\STATE (b). ${\mathbf{S}_{{i}}}=\beta_{1}*f_x(\mathbf{X}_{{i}})$
		\STATE (c). $L={\sum _{i=1}^{k}{l\left({{\mathbf{S}_{{i}}},{\mathbf{Y}_{{i}}}}\right)}}$
		\STATE (d). Get ${{h_2(\mathbf{Z}_1)}}$ according to equation (\ref{equ:f2star}).
		\STATE (e). $\beta_{1}^{\star}={{U_1}}({{h}_{{2}}}(\mathbf{Z}_1),
		{{\alpha}_{{1}}})$
		\ENDFOR
		
		\STATE 3. Train the update branch by minimizing the loss.
		\FOR{$i \in {0,\hdots,k}$}
		\STATE (a). ${\mathbf{S}_{{i}}}^{\star}=\beta^{\star}_{1}*f_x(\mathbf{X}_{{i}})$
		\STATE (b). $L^{\star}={\sum _{i=1}^{k}{l\left({{\mathbf{S}_{{i}}}^{\star},{\mathbf{Y}_{{i}}}}\right)}}$
		\STATE (c). Minimize $L^{\star}$ to update ${{\alpha}^{0}}$ by SGD.
		\ENDFOR
		\ENDWHILE
	\end{algorithmic}
\end{algorithm}

\subsection{Online Tracking}\label{sec::tracking}
After offline training, the update branch is totally fixed and used for initialization and update during online testing .

\noindent
{{\bf{Initialization. }}}
Given the ground truth in the first frame, we crop a target patch ${\mathbf{Z}_{{1}}}$ and
a search region ${\mathbf{X}_{{1}}}$ as inputs of the network.
Then, we can obtain the optimal template ${{\beta}^{{\star}}}$ according to equations (\ref{equof2}$-$\ref{equ:final}).
Besides, the updated target features ${{h}_{{2}}}(\mathbf{Z_1})$ is calculated through equation (\ref{equ:f2star}) and used to update the template in the following frames.

\noindent
{{\bf{Online Update. }}}
We update the template ${{\beta}^{{\star}}}$ with one reliable training sample through one iteration.
We save the reliable sample $({\mathbf{X}_{{i}}},{\mathbf{Y}_{{i}}})$ according
to tracking results and use it to update the current template ${{\beta}^{{\star}}}$ based on equations (\ref{equof2}$-$\ref{equ:final})
( replacing $f_2(\mathbf{Z})$, $\mathbf{X}$, $\mathbf{Y}$ with ${{h}_{{2}}}(\mathbf{Z})$, $\mathbf{X}_i$, $\mathbf{Y}_i$ ).
Namely, we obtain updated feature ${{h}_{{2}}}(\mathbf{Z}_1)$ through the initial frame. Then, the update branch of network is used to update ${{h}_{{2}}}(\mathbf{Z}_1)$ according to the reliable sample $({\mathbf{X}_{{i}}},{\mathbf{Y}_{{i}}})$ and produce optimal templates ${{\beta}^{\star}}$ for the regression part.

\subsection{Implementation Details}\label{sec::detail}
The feature extraction $f_x(.)$ for the search region consists of five convolutional layers with the same structure and parameters as SiameseFC~\cite{Bertinetto-ECCV16-SiamesFC}.
The shallow target features $f_2(.)$ are from the second convolutional layers of SiameseFC.
There are three convolutional layers in $U_1$ which have the same structure with the last three layers of SiameseFC.
The kernel size of the convolutional layer in $U_2$ is $3 \times 3$.
%
%
The size of template $\beta$ and $\beta^{\star}$ is $6 \times 6$ and the size of score map is $17 \times 17$.
During tracking, we update the template ${{\beta}^{{\star}}}$ every 5 frames.
The reliable training sample is chosen according to the max value of the score map. We set the max value of the score map in the first frame as a threshold $thre$. If the max value of the current score map is larger than $thre*0.5$, we think that the result is accurate and crop the training sample ${\mathbf{X}_{{t}}}$ as the reliable training sample.
The scale evaluate, learning rate and training epoch in the proposed method are the same as those in SiameseFC~\cite{Bertinetto-ECCV16-SiamesFC}.
To take the trade-off between the fast adaptation and error accumulation, the final template is obtained by combining the initial template and $\beta^{\star}$.
We only train the network on ILSVRC2014 VID dataset and the whole network is fixed during inference.

\begin{figure}[t]
	\centering
	\begin{tabular}{c@{}c}
		\includegraphics[width=0.495\linewidth]{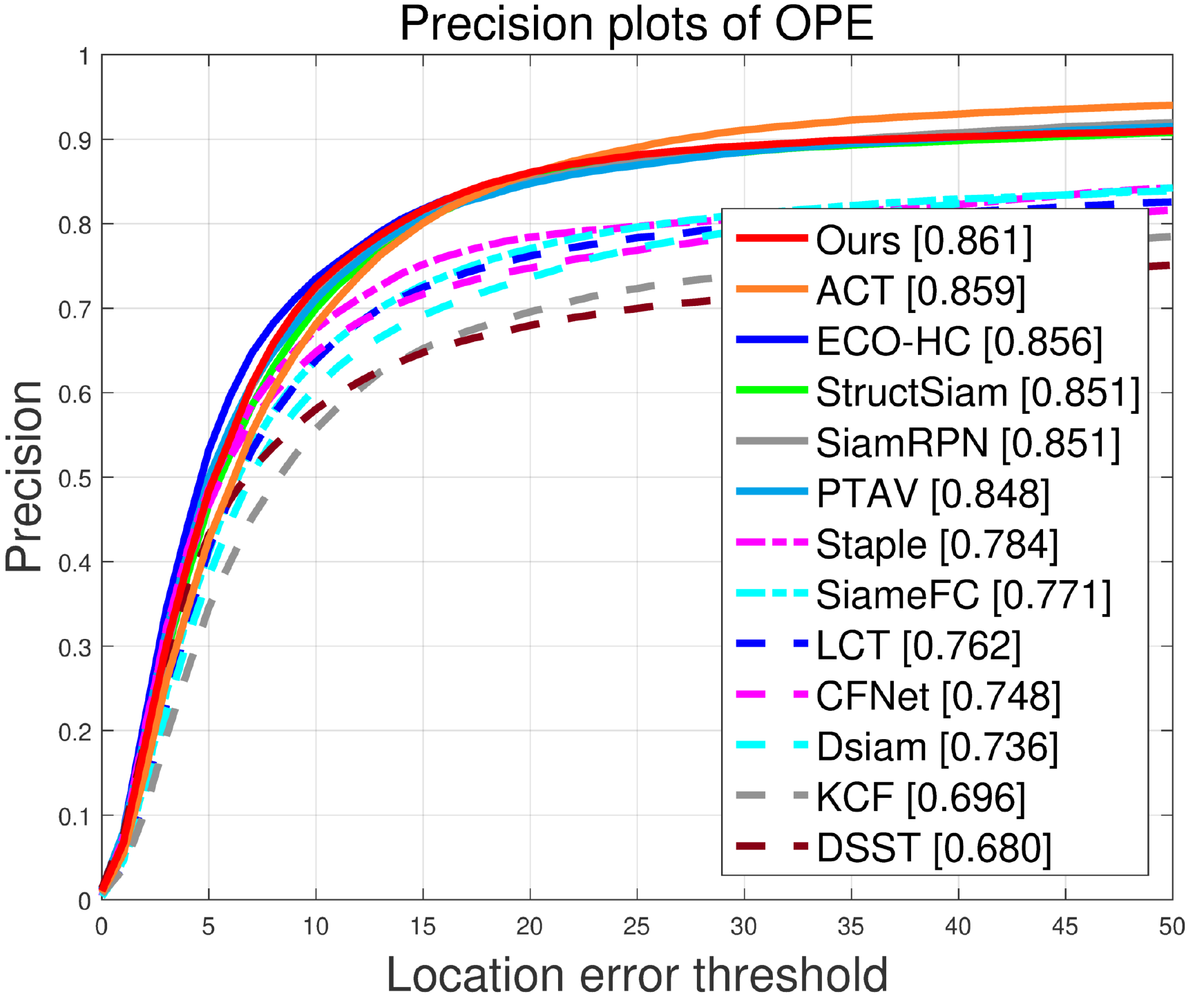}  &
		\includegraphics[width=0.495\linewidth]{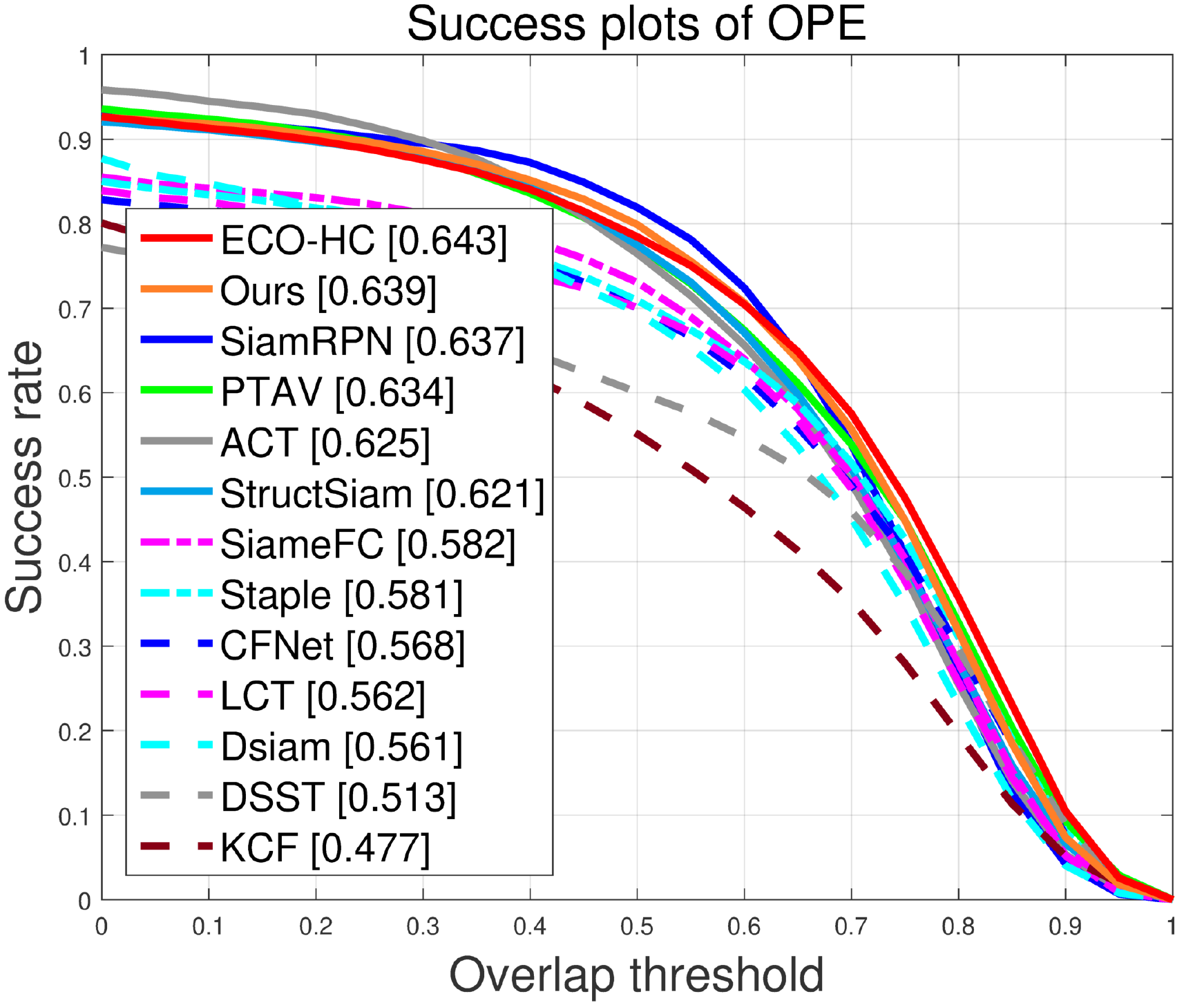} \\
	\end{tabular}
	\caption{Precision and success plots on the OTB-2015 dataset.}
	\label{fig-otb}
\end{figure}
\begin{figure}[t]
	\centering
	\begin{tabular}{c@{}c}
		\includegraphics[width=0.495\linewidth]{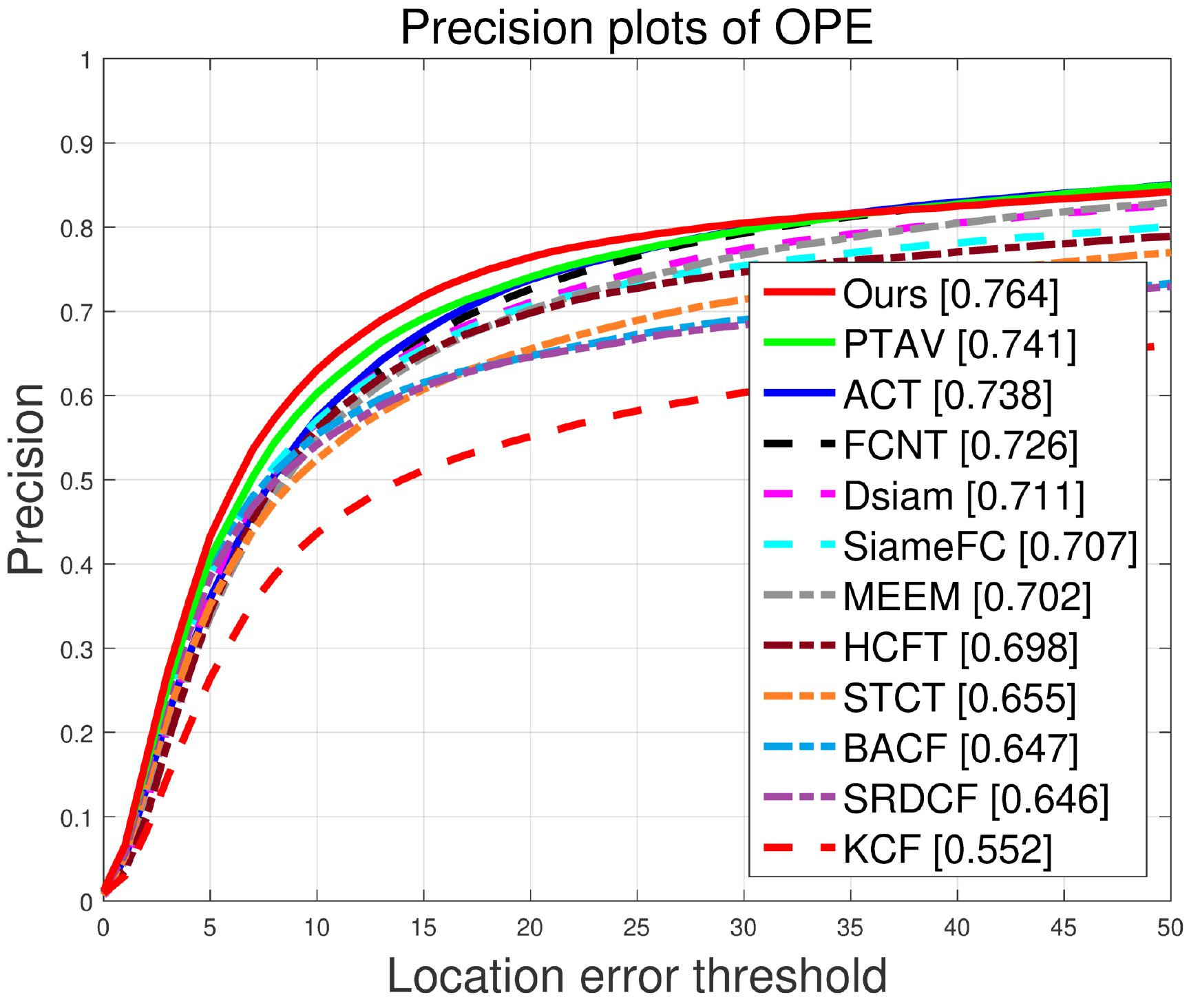}    &
		\includegraphics[width=0.495\linewidth]{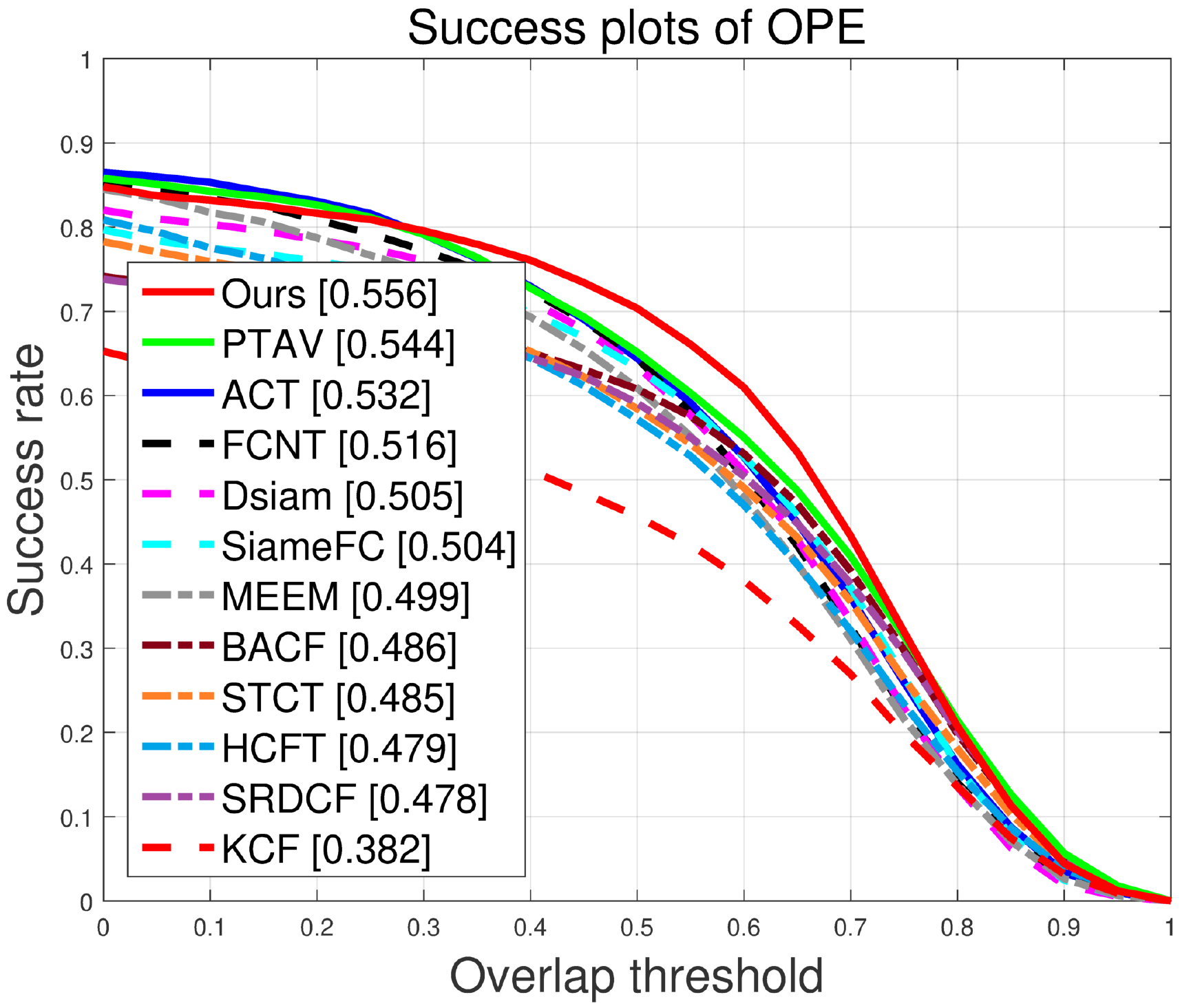}   \\
	\end{tabular}
	\caption{Precision and success plots on the TC128 dataset.}
	\label{fig-tc128}
\end{figure}
\begin{figure}[t]
	\centering
	\begin{tabular}{c@{}c}
		\includegraphics[width=0.495\linewidth]{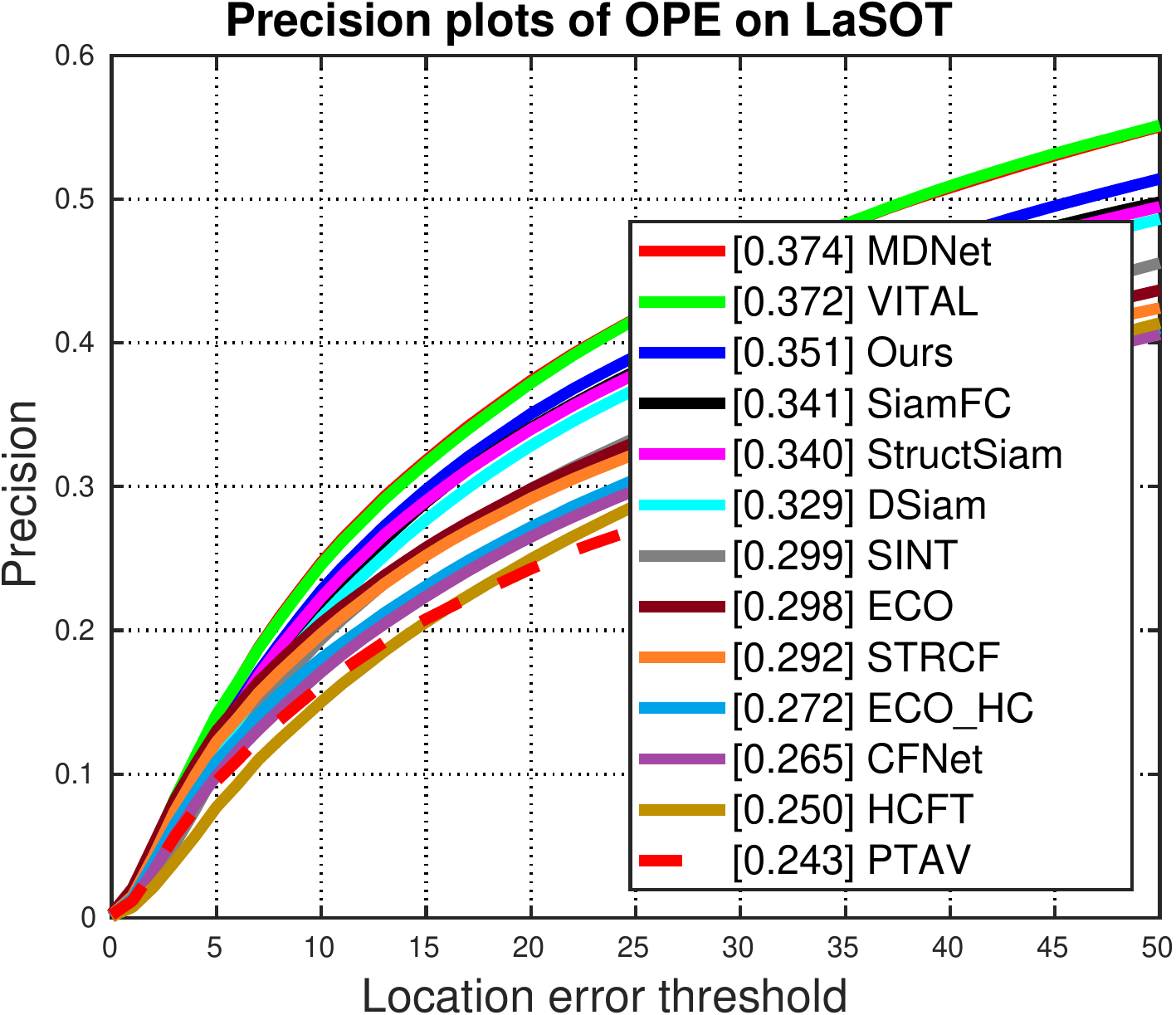}    &
		\includegraphics[width=0.495\linewidth]{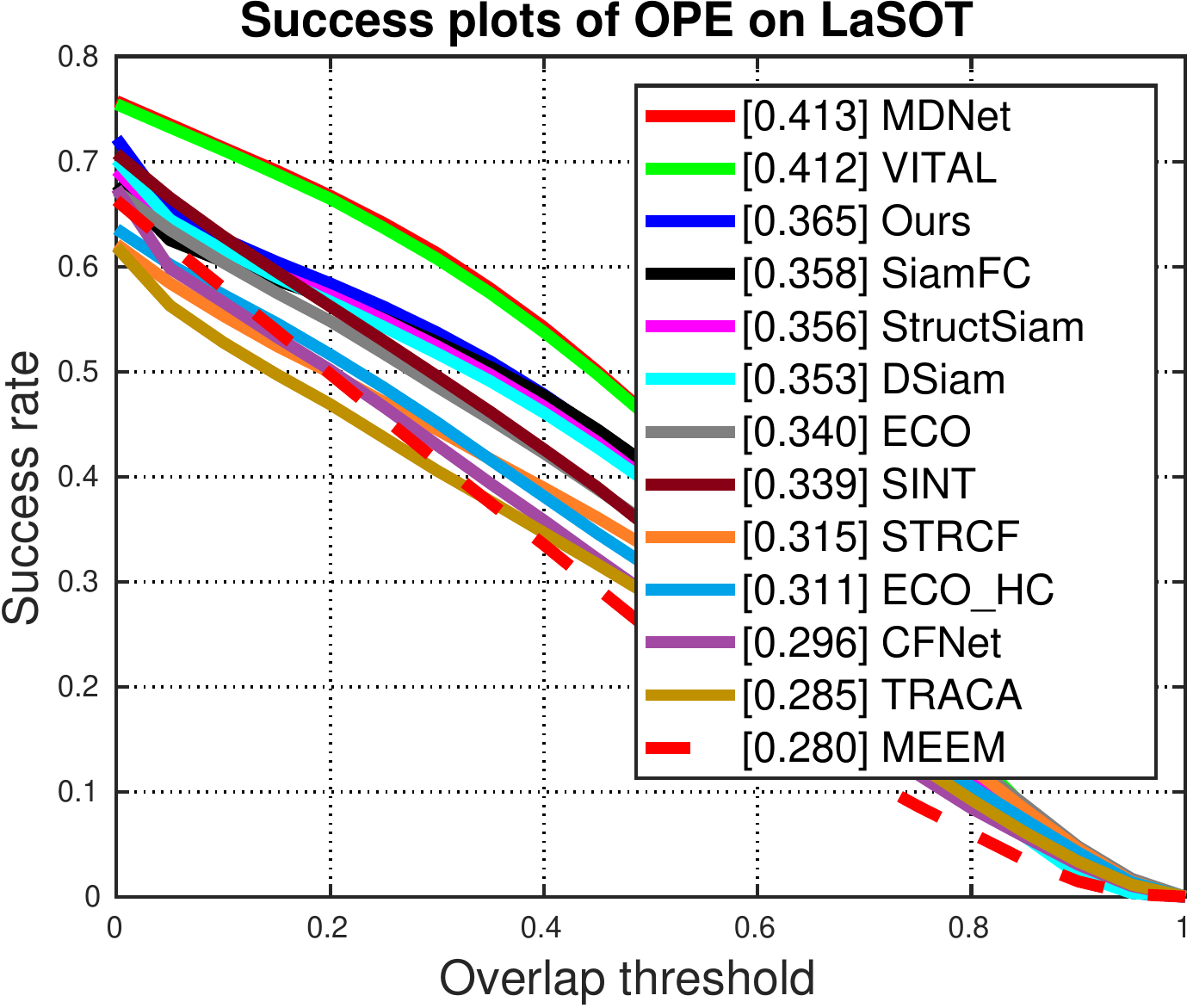}   \\
	\end{tabular}
	\caption{Precision and success plots on the LaSOT dataset.}
	\label{fig-lasot}
\end{figure}

\vspace{-1.5mm}
\section{Experiments}
Our tracker is implemented in Python with the Pytorch framework, which runs at $\mathbf{80}$\emph{fps}
with an intel i7 3.2GHz CPU with 32G memory and a Nvidia 1080ti GPU with 11G memory.
We compare our tracker with many state-of-the-art trackers with real-time performance
(i.e., their speeds are faster than $25$\emph{fps}) on recent benchmarks,
including OTB-2015~\cite{WuLY-15paim-OTB100}, TC-128~\cite{TC128},
VOT-2017~\cite{KristanLMFPCVHL-ICCVW17-VOT17} and LaSOT~\cite{LaSOT}.

\begin{figure*}[t]
	\footnotesize
	\begin{center}
		\includegraphics[width=0.9\linewidth]{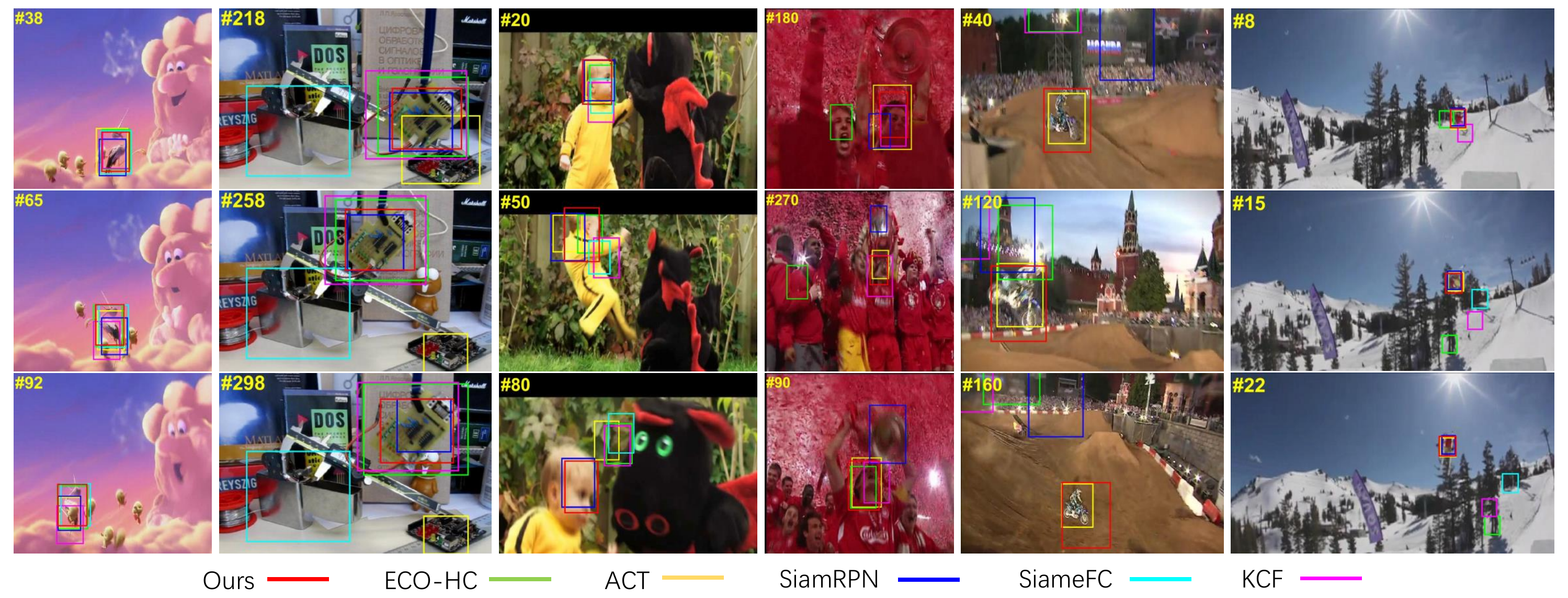}   \\
	\end{center}
 	\vspace{-3mm}
	\caption{Representative visual results of different tracking algorithms on the OTB-2015 dataset.}
	\label{fig-visual}
\end{figure*}


\subsection{Evaluation on the OTB-2015 dataset}
The OTB-2015~\cite{WuLY-15paim-OTB100} dataset is one of the most popular
benchmarks, which consists of $100$ challenging video clips annotated with $11$
different attributes. We refer the reader to ~\cite{WuLY-15paim-OTB100} for more detailed information.
Here, we adopt both
success and precision plots to evaluate different trackers on OTB-2015.
The precision plot reports the percentages that the center location errors
are smaller than certain thresholds. Whereas the success plot reports the
percentages of frames where the overlap between the predicted and the
ground truth bounding boxes is higher than a series of given ratios.
We compare our algorithm with twelve state-of-the-art trackers including nine
real-time deep trackers (ACT~\cite{chen-ECCV18-act},
StructSiam~\cite{zhang-eccv2018-structsiam}, SiamRPN~\cite{Li-2018CVPR-siameserpn},
ECO-HC~\cite{Martin-ECO-CVPR17}, PTAV~\cite{ptav-iccv17},
CFNet~\cite{cfnet-cvpr17}, Dsiam~\cite{Guo-dsiam-iccv2017}, LCT~\cite{LCT15},
SiameFC~\cite{Bertinetto-ECCV16-SiamesFC}) and three traditional trackers
(Staple~\cite{staple16}, DSST~\cite{Danelljan-BMVC14-DSST},
KCF~\cite{Henriques-TPAMI15-KCF}).

\begin{table}[t]
	\centering
	\caption{The accuracy (A), robustness (R) and expected average overlap
		(EAO) scores of different trackers on VOT2017.}
	\vspace{1mm}
	\setlength{\tabcolsep}{5mm}{
		\begin{tabular}{l||c|c|c}
			\hline {Trackers} & {\textbf{A}} & {\textbf{R}} & {\textbf{EAO}} \\
			\hline
			\hline {Ours} &\textcolor{green}{0.507}  & \textcolor{red}{0.375} & \textcolor{red}{0.247} \\
			\hline {SiamRPN} & 0.490 & \textcolor{green}{0.460} & \textcolor{blue}{0.244} \\
			\hline {CSRDCF++} & 0.459 & \textcolor{blue}{0.398} & \textcolor{green}{0.212} \\
			\hline {SiamFC} & 0.502 & 0.604 & 0.182 \\
			\hline {ECO\_HC} & 0.494 & 0.571 & 0.177 \\
			\hline {Staple} & \textcolor{blue}{0.530} & 0.688 & 0.170 \\
			\hline {KFebT} & 0.451 & 0.684 & 0.169 \\
			\hline {SSKCF} & \textcolor{red}{0.530} & 0.656 & 0.164 \\
			\hline {CSRDCFf} & 0.475 & 0.646 & 0.158 \\
			\hline {UCT} & 0.490 & 0.777 & 0.145 \\
			\hline {MOSSEca} & 0.400 & 0.810 & 0.139 \\
			\hline {SiamDCF} & 0.503 & 0.988 & 0.135 \\
			\hline
		\end{tabular}}
		\label{vot-table}
	\end{table}

Figure~\ref{fig-otb} illustrates the precision and success plots of all compared trackers
over OTB-2015, which shows the proposed tracker achieves very good performance
(merely a slightly lower than ECO-HC in success).
Especially, our tracker performs significantly better than the baseline model
(SiameseFC) by almost 8$\%$ in precision and 6$\%$ in success.
To facilitate more detailed analysis, we demonstrate the visual
results of some representative methods in Figure~\ref{fig-visual}.
From these figures, we can see that our method can well handle various challenging
factors and consistently achieve good performance.

\subsection{Evaluation on the TC-128 dataset}
The TC128~\cite{TC128} dataset consists of $128$ fully-annotated image sequences
with $11$ various challenging factors, which is larger than OTB-2015 and focuses
more on color information. We also adopt both success and precision plots to evaluate
different trackers (the same evaluation protocol as OTB-2015). We compare our algorithm
with eleven trackers, including ACT~\cite{chen-ECCV18-act}, PTAV~\cite{ptav-iccv17},
Dsiam~\cite{Guo-dsiam-iccv2017},
SiameFC~\cite{Bertinetto-ECCV16-SiamesFC}, HCFT~\cite{Ma-ICCV15-HCFT},
FCNT~\cite{Wang-ICCV15-FCNT}, STCT\cite{Wang-CVPR16-STCT},
BACF~\cite{Galoogahi-iccv17-bacf}, SRDCF~\cite{Danelljan-ICCV15-SRDCF},
KCF~\cite{Henriques-TPAMI15-KCF} and MEEM~\cite{Zhang-ECCV14-MEEM}.
Figure~\ref{fig-tc128} shows that our tracker achieves the best results in terms of both precision and success criterion.

\subsection{Evaluation on the VOT2017 dataset}
The VOT2017~\cite{KristanLMFPCVHL-ICCVW17-VOT17} dataset contains
$60$ short sequences annotated with $6$ different attributes.
%
According to its evaluation protocol, the tested tracker is re-initialized whenever
a tracking failure
is detected. In this benchmark, the accuracy (A) and robustness
(R) as well as expected average overlap (EAO) are three important criterion.
Different trackers are ranked based on the EAO criterion.
We refer the reader to~\cite{KristanLMFPCVHL-ICCVW17-VOT17} for more
detailed information.
In this subsection, we compare our algorithm with top ten trackers reported
in the VOT2017 real-time Challenge~\cite{KristanLMFPCVHL-ICCVW17-VOT17} and another state-of-the-art tracker SiamRPN~\cite{Li-2018CVPR-siameserpn}.
Table~\ref{vot-table} shows that our tracker achieves the best performance in terms of
EAO while maintaining a very competitive accuracy and robustness.
The EAO of our tracker is higher than the winner (CSRDCF++) of the VOT2017 real-time Challenge by $3.5\%$.
Our tracker can also perform better than SiamRPN whose training data (over 100,000 videos) is much larger than ours (about 4,000 videos).

\subsection{Evaluation on the LaSOT dataset}
The LaSOT~\cite{LaSOT} dataset is a very large-scale dataset consisting of
$1,400$ sequences with $70$ categories and more than $3.5$M frames in total.
The average frame length of this dataset is more than $2,500$ frames. Up to
now, this dataset is the largest for visual tracking.
Following one-pass evaluation, different trackers are compared based on three
criteria including precision, normalized precision and success.
We also adopt precision and success plots to compare $35$ trackers and show
the performance of the top $12$ trackers in Figure~\ref{fig-lasot} (more compared results are
presented in the supplementary material).
From Figure~\ref{fig-lasot}, we can see that our tracker performs the third-best
in this dataset. Although MDNet and VITAL achieve better accuracies than our tracking algorithm, their speeds are far from the real-time requirement (MDNet, \textsl{1fps} and VITAL, \textsl{1.5fps}).

\subsection{Ablation Analysis}
\label{exp:ablation}
{\flushleft{\bf{Self-comparison.}}}
To verify the contribution of each component in our algorithm, we implement and evaluate several variations of our approach (Ours) on OTB-2015. These versions include: (1) `Ours w/o M': GradNet without template generalization training process; (2) `Ours w/o MG': GradNet removed template generalization training process and gradient application. It can be seen as SiameseFC with two unshared branches; (3) `Ours w/o U': the proposed method without template update;  (4) `Ours w 2U': the two sub-nets (in purple) in Figure~\ref{figs1} do not share parameters; (5) `Ours-baseline': SiameseFC.

\begin{table}[h]
	\centering
	\caption{Precision and success scores on OTB-2015 for different variations of our algorithm. }
	\centering
	\setlength{\tabcolsep}{5mm}{
		\begin{tabular}{|l||c|c|c|}
			\hline
			{Variations} & {\textbf{PRE}} & {\textbf{IOU}} & {\textbf{FPS}} \\
			\hline
			\hline
			{Ours} & 0.861 & 0.639 & 80  \\
			\hline
			{Ours w/o M} & 0.823 & 0.615 & 80  \\
			\hline
			{Ours w/o MG} & 0.717 & 0.524 & 94 \\
			\hline
			{Ours w/o U} & 0.775 & 0.552 & 85 \\
			\hline
			{Ours w 2U} & 0.833 & 0.628 & 80 \\
			\hline
			{Ours-baseline} & 0.771 & 0.582 & 94  \\
			\hline
		\end{tabular}}
		\label{table:db}
	\end{table}

The performance of all variations and our final method is reported in Table~\ref{table:db},  from which we can see that all components facilitate improving the tracking accuracy. For examples, the comparison of the `Ours w/o M' and final methods demonstrates the template generalization training method could effectively learn an expected GradNet.
With the same amount of training data,  `Ours' improves the precision and IOU score of `Ours-baseline' about $9\%$ and $5\%$ respectively, which demonstrates the effectiveness of the GradNet.

\begin{figure}[t]
	\centering
	\includegraphics[width=1\linewidth]{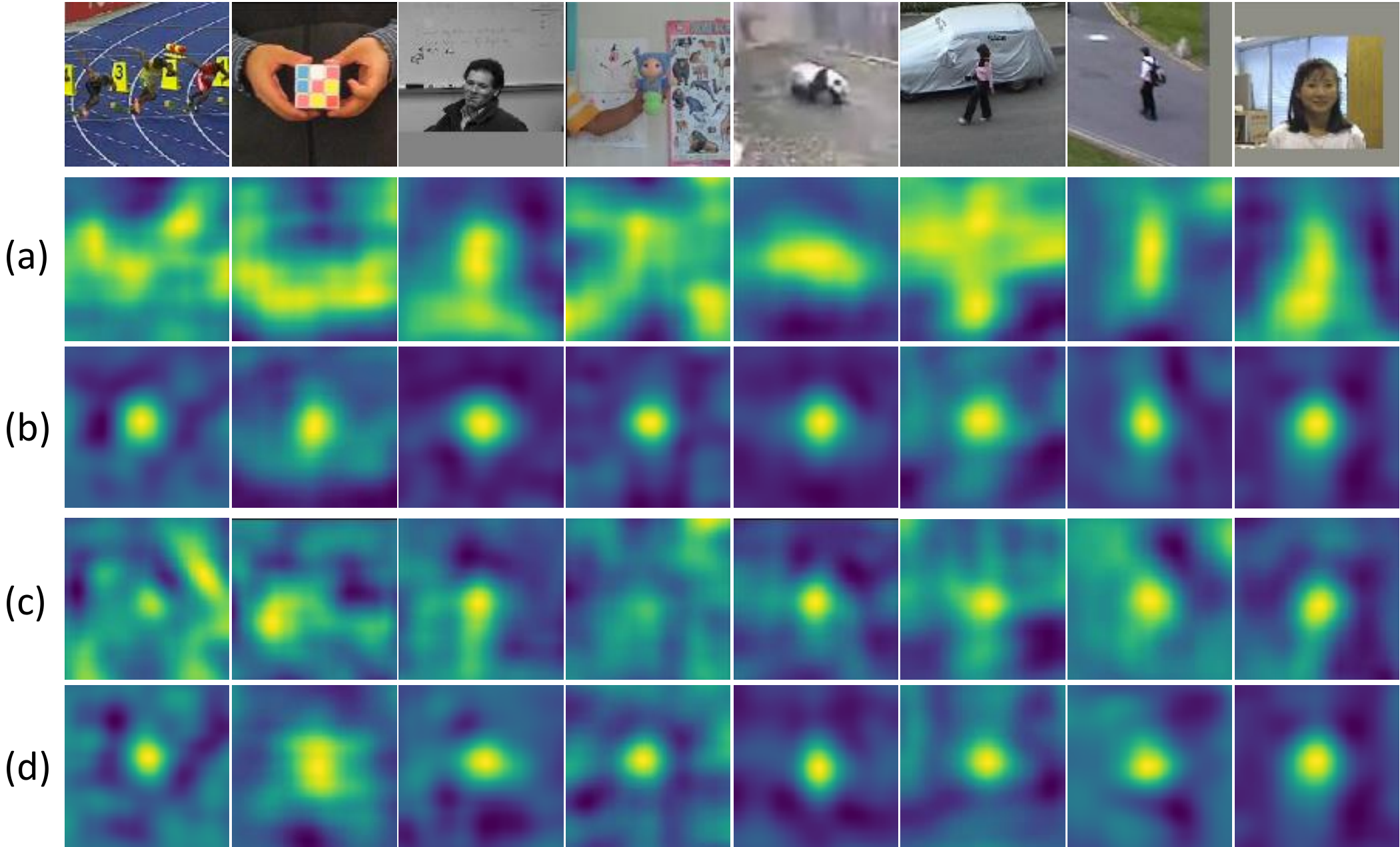}\\
	\caption{The first row displays the search regions from different videos. (a) and (b) shows $\mathbf{S}$ and $\mathbf{S}^{\star}$ of our model; (c) and (d) shows  $\mathbf{S}$ and $\mathbf{S}^{\star}$ of the model without template generalization. Model through template generalization can get general initial score maps $\mathbf{S}$ and optimal final score maps $\mathbf{S}^{\star}$.} \label{fig:s1s2}
\end{figure}
{\flushleft{\bf{Training Analysis.}}}
To further analyze the template generalization, we show the initial score map $\mathbf{S}$ and the optimal score map $\mathbf{S}^{\star}$ of two different training methods in Figure~\ref{fig:s1s2}.
The initial score maps of the model with template generalization (a) are noisy score maps where the approximate area of all objects has high response values.
After the template updating based on gradients, the promising score maps (b) only have a high response at the target position.
Differently, the model without template generalization is likely to output initial score maps (c) with a high response at the target position directly.
Thus, we think the model trained by template generalization learns different tasks in the initial embedding and template update processes. During initial embedding, it learns a general template to detect the target and background clutter. This manner provides the model more discriminative gradients. Then, the model learns to update the template based on these gradients in the template update process.
The discriminative gradients enable the fast adaptation of the network.

\section{Conclusions}
In this work, we propose a GradNet for template update, achieving accurate tracking with a high speed.
The two sub-nets in GradNet exploits the discriminative information in gradients through feed-forward and backward operations and speeds up the hand-designed optimization process.
To take full use of gradients and obtain versatile templates, a template generalization method is applied during offline training, which can force the update branch to concentrate on the gradient and avoid overfitting.
Experiments on four benchmarks show that our method
significantly improves the tracking performance compared with other real-time trackers.

\vspace{-3mm}
{\flushleft\textbf{Acknowledgements}}
The paper is supported in part by Natioal Natural Science Foundation of China No.61725202, 61829102, 61751212 and the Fundamental Research Funds for the Central Universities under Grant Nos. DUT19GJ201, DUT18JC30.

\clearpage
{\small
\bibliographystyle{ieee_fullname}
\bibliography{egbib}
}

\end{document}